\documentclass[conference]{IEEEtran}

\usepackage{cite}
\usepackage{color}
\usepackage{graphicx}
\usepackage{epstopdf}
\usepackage{amsmath}
\usepackage{url}

\DeclareMathOperator*{\argmax}{arg\,max}
\begin{document}

\title{
\begin{normalsize}
\begin{flushleft}
\copyright 2017 IEEE. Personal use of this material is permitted. Permission from IEEE must be obtained for all other uses, in any current or future media, including reprinting/republishing this material for advertising or promotional purposes, creating new collective works, for resale or redistribution to servers or lists, or reuse of any copyrighted component of this work in other works.\\
\medskip
This is a preprint version. The final version of this paper is available at \url{http://ieeexplore.ieee.org/document/7966389} (DOI:10.1109/IJCNN.2017.7966389).\\
\end{flushleft}
\hrulefill
\color{white}
.\\
\end{normalsize}
\color{black}
Batch Reinforcement Learning on the Industrial Benchmark: First Experiences}

\author{\IEEEauthorblockN{Daniel Hein\IEEEauthorrefmark{1}\IEEEauthorrefmark{2},
Steffen Udluft\IEEEauthorrefmark{2},
Michel Tokic\IEEEauthorrefmark{2}, 
Alexander Hentschel\IEEEauthorrefmark{2},
Thomas A. Runkler\IEEEauthorrefmark{1}\IEEEauthorrefmark{2} and
Volkmar Sterzing\IEEEauthorrefmark{2}}
\IEEEauthorblockA{\IEEEauthorrefmark{1}Technical University of Munich, Department of Informatics, Boltzmannstr. 3, 85748 Garching, Germany}
\IEEEauthorblockA{\IEEEauthorrefmark{2}Siemens AG, Corporate Technology, Otto-Hahn-Ring 6, 81739 Munich, Germany}
}

\maketitle

\begin{abstract}
The Particle Swarm Optimization Policy (PSO-P) has been recently introduced and proven to produce remarkable results on interacting with academic reinforcement learning benchmarks in an off-policy, batch-based setting. 
To further investigate the properties and feasibility on real-world applications, this paper investigates PSO-P on the so-called Industrial Benchmark (IB), a novel reinforcement learning (RL) benchmark that aims at being realistic by including a variety of aspects found in industrial applications, like continuous state and action spaces, a high dimensional, partially observable state space, delayed effects, and complex stochasticity.

The experimental results of PSO-P on IB are compared to results of closed-form control policies derived from the model-based Recurrent Control Neural Network (RCNN) and the model-free Neural Fitted Q-Iteration (NFQ). 

Experiments show that PSO-P is not only of interest for academic benchmarks, but also for real-world industrial applications, since it also yielded the best performing policy in our IB setting.
Compared to other well established RL techniques, PSO-P produced outstanding results in performance and robustness, requiring only a relatively low amount of effort in finding adequate parameters or making complex design decisions.
\end{abstract}

\section{Introduction}
The process of controlling industrial plants, or parts of such, involves a variety of challenging aspects that reinforcement learning (RL) \cite{sutton_reinforcement_1998} algorithms need to tackle. 
For coping adequately with the complexity of real-world systems, important challenges that need to be considered are: 
continuous state and action spaces, high-dimensional and only partially observable state spaces, stochasticity induced from heteroscedastic sensor noise and latent variables, delayed effects, multi-criterial reward components, and non-stationarity in the optimal steerings, i.e.~the optimal policy will not approach a fixed operation point.

Here, we consider applications where on-line learning, like the classical temporal-difference learning approach \cite{sutton1988learning}, is prohibited for safety reasons, since it requires exploration on the plant's dynamics.
In contrast, batch RL algorithms generate a policy based on existing data, which is deployed to the plant after training.
In this setting, either the value function or the system dynamics are trained by historic operational plant data, which is a set of four-tuples \textit{(observation, action, reward, next observation)} called \textit{batch} in the following.
Research from the past two decades suggests that the family of batch RL algorithms \cite{gordon:95,OrmoneitSen2002,LagoudakisParr_2003,ernst_treebased_2005}, meet the requirements of real-world systems, especially when involving neural networks modeling either the state/action value function \cite{riedmiller:051,riedmiller:05,nrr:07,pgnrr:07,Riedmiller2009}, or the system dynamics \cite{Bakker:2004,schafer2008reinforcement,Depeweg2016}.
Moreover, batch RL algorithms are data efficient \cite{riedmiller:051,schaefer2007recurrent}, because the batch data is utilized repeatedly during the training phase.

In the following we investigate different RL approaches on the \textit{Industrial Benchmark} (IB) \cite{hein:162} that comes with challenges being vital in industrial settings as described above. 
We report on results for applying \textit{Particle Swarm Optimization Policy} (PSO-P) \cite{hein:16}, which is a powerful algorithm for RL with continuous actions that achieves remarkable results out of the box.
The actions to perform are derived from rollouts on a system model, which simulates the IB's transition dynamics.
This model, a recurrent neural network, was trained on a batch of transitions sampled from applying random actions to the IB. 
In real-world applications, however, transitions are usually available in form of historic operational data, that might have been produced by a constant default controller. 

We compare the performance of PSO-P on IB with two other RL approaches, that utilize the batch in different ways. 
First, the \textit{Recurrent Control Neural Network} (RCNN) \cite{schaefer2007recurrent}, which is a model-based RL algorithm for continuous actions, that uses the system model during training of the controller. 
Second, \textit{Neural Fitted Q-Iteration} (NFQ) \cite{riedmiller:051}, a model-free RL algorithm for discrete actions, where the controller is learned via iteratively applying Watkins' Q-learning algorithm \cite{Watkins:1989} on the batch data. 
Here, the system model is used to select the best policy after the NFQ training has finished. 
Policy selection is necessary, because the performance of NFQ policies can decrease significantly over iterations. 
This is a well-known phenomenon in the context of neuro-dynamic programming \cite{gabel:06}.

\section{Industrial Benchmark}

The Industrial Benchmark\footnote{Source code available at: \url{http://github.com/siemens/industrialbenchmark}} (IB) \cite{hein:162} aims at being realistic in the sense that it includes a variety of aspects that we found to be vital in industrial applications. 
It is not designed to be an approximation of any specific real-world system, but to pose a comparable hardness and complexity found in many industrial applications. 
State and action space are continuous, the state space is high-dimensional and only partially observable. 
The actions consist of three continuous components and affect three steerings. 
Moreover, the IB includes stochastic and delayed effects.
The optimization task is multi-criterial in the sense that there are two reward components that show opposite dependencies on the actions. 
The dynamical behavior is heteroscedastic with state-dependent observation noise and state-dependent probability distributions, based on latent variables. 
Furthermore, it depends on an external driver that cannot be influenced by the actions. 
The IB is designed such that the optimal policy will not approach a fixed operation point in the three steerings. 
Any specific choice is driven by our experience with industrial challenges.

At any time step $t$ the RL agent can influence the environment (IB) via actions $a_t$ that are three dimensional vectors in $[-1,1]^3$. 
Each action can be interpreted as three proposed changes to three observable state variables called \textit{current steerings}. 
Those current steerings are: velocity $v$, gain $g$, and shift $h$. 
Each of those is limited to $[0,100]$, yielding
\begin{align}
	a_t &= (\Delta v,\Delta g, \Delta h), \\
	v_{t+1} &= \max(0,\min(100,v+d_v\Delta v)), \\
	g_{t+1} &= \max(0,\min(100,g+d_g\Delta g)), \\
	h_{t+1} &= \max(0,\min(100,h+d_h\Delta h)),
\end{align}
with scaling factors $d_v=1$, $d_g=10$, and $d_h=5.75$.

After applying the action $a_t$, the environment transitions to the next time step $t+1$, yielding the internal state $s_{t+1}$. 
State $s_{t}$ and successor state $s_{t+1}$ are the Markovian states of the environment. 
In many industrial applications, an operator-defined load $p_t$ is applied to the system. 
Depending on load $p_t$ and the control values $a_t$, the system shows fatigue $f_t$ and consumes resources such as power, fuel, etc., represented by consumption $c_t$.
Both, $p_t$ and $a_t$, are external drivers for the IB. 
In response, the IB generates values for $c_{t+1}$ and $f_{t+1}$, which are part of the internal state $s_{t+1}$.
The reward is solely determined by $s_{t+1}$:
\begin{equation}
r_t =-c_{t+1}-3f_{t+1}.
\label{eq:reward}
\end{equation}
In the real world tasks that motivated the IB, the reward function has always been known explicitly. 
That is why we assume that here the reward function is also known and consumption and fatigue are observable.
However, except for the values of the steerings, the remaining part of the Markov state remains unobservable. This yields an observation vector $o_t \subset s_t$ consisting of:
\begin{enumerate}
	\item the current steerings: velocity $v_t$, gain $g_t$, and shift $h_t$,
	\item the external driver: set point $p_t$,
	\item and the reward relevant variables: consumption $c_t$ and fatigue $f_t$.
\end{enumerate}

One of the IB's features is the possibility to freeze its stochasticity. 
On the one hand, for data generation, online RL experiments, and policy evaluation, stochasticity makes the benchmark realistic and challenging. 
On the other hand, there are some settings where freezing the randomness in the stochastic effects might be useful.
This is realized by remembering the applied seed of the IB-internal pseudo-random number generator (RNG).
For instance, in the experiments presented in Section \ref{experiments}, we searched for the true maximum reward given a certain Markov state and its current RNG seed, as the upper bound for the RL technique performance evaluation.
This has been done by applying PSO-P directly on the IB system dynamics, provided with full knowledge about the future, encoded in the RNG seed.

\section{PSO-P}

In the Particle Swarm Optimization Policy (PSO-P) framework \cite{hein:16}, solving an RL problem is reformulated as an optimization problem. 
RL is an area of machine learning, where the Markov decision problem has to be solved by learning from observed state transitions $(s_{t},a_{t})\rightarrow(s_{t+1},r_{t})$, with $s_t$ and $s_{t+1}$ representing the Markovian states, $a_t$ the applied action, and $r_t$ the real-valued reward, at discrete time steps $t$ and $t+1$. The goal is to find a policy maximizing the expected cumulative reward $\mathcal{R}=\sum^{\infty}_{k=0}\gamma^k r_k$, called \textit{return}, where $0\leq\gamma\leq1$ is the so-called discount factor \cite{sutton_reinforcement_1998}.

Since the true underlying Markovian state $s_t$ is not observable in the IB, it is approximated by a sufficient amount of historic observations, i.e.~the information contained in $s_t$ is approximated by $(o_{t-H},\ldots,o_t)$ with horizon $H$.

Given a system model $m(o_{t-H},\ldots,o_{t},a_{t})=(o_{t+1},r_{t})$ (see Section~\ref{experiments}), trained by supervised learning methods on previous observations, finding the best action $a_t$, for a given an observation horizon with respect to the return $\mathcal{R}$, is described as
\begin{align}
    \mathcal R(o_{t-H},\ldots,o_{t},\mathbf{x})&=\sum^{T-1}_{k=0}\gamma^kr_{t+k}, \text{with}\\
    \nonumber (o_{t+k+1},r_{t+k})&=m(o_{t+k-H},\ldots,o_{t+k},a_{t+k}).
\end{align}
The discount factor $\gamma$ is chosen such that at the end of the time horizon $T$, the last reward accounted for is weighted by $q\in[0,1]$, computed by $\gamma=q^{1/(T-1)}$.

Particle swarm optimization \cite{kennedy:95} (PSO) is then searching for the optimal action sequence $\hat{\mathbf{x}}=(a_t,a_{t+1},\ldots,a_{t+T-1})$ satisfying
\begin{equation}
	\hat{\mathbf{x}}\in\argmax_{\mathbf{x}\in\mathcal{A}^T}\mathcal R (o_t,\mathbf{x}).
\end{equation}

Analogues to receding horizon control (RHC), only the first element of $\hat{\mathbf{x}}$ is returned. 
This yields an RL policy $a_t=\pi(o_{t})$, which conducts an optimization for every new observation.
This might be computationally expensive, but it does not rely on a predefined closed-form policy structure, which very often is a hard to asses {\it a priori} assumption for common RL algorithms on novel problems.

\section{Experiments}
\label{experiments}

We compare the performance of PSO-P on IB with the well established RCNN and NFQ methods. 
For all of the applied RL techniques, we required an adequate system model $m$ simulating IB trajectory rollouts:
\begin{itemize}
	\item RCNN: The RCNN is trained on $m$ to calculate the respective gradients for the policy's weight update step.
	\item NFQ: Despite NFQ being considered model-free RL, it is still very useful to evaluate the policy's performance after each Q-iteration step on $m$, since performance drops during the training are very likely to occur when applying NFQ on off-policy batch data. Therefore, in our experiments the policy with the best performance according to model $m$ is saved and returned as the final NFQ training result. 
	\item PSO-P: The policy represented by PSO-P utilizes $m$ during runtime. PSO-generated trajectories are rated by performing rollouts on $m$ in every policy query.
\end{itemize}

In the following experiments the system model $m$ predicts consumption $c$ and fatigue $f$ for each step of the rollout, and the reward is computed according to Eq.~\eqref{eq:reward}.

To generate an adequate training data set $\mathcal{D}$, we initialized the IB 10 times for each set point $p=\{10,20,\ldots,100\}$ and produced random trajectories of lengths $1\,000$, resulting in $|\mathcal{D}|=100\,000$.

The system model $m$ was chosen to consist of two recurrent neural networks (RNN) $m_c$ and $m_f$, to predict consumption $c$ and fatigue $f$, respectively.
Both models are unfolded a sufficient number of steps into the past ($H_c=30$ time steps for $m_c$, $H_f=10$ time steps for $m_f$) and 50 time steps into the future. 
In each time step they take the observable variables of the past and present as input. 
Whereas, in the future branch of the RNNs only the steerings (velocity, gain, and shift) are used as input. 
The topology of these RNNs is described in \cite{Duell:2012b} as "Markov Decision Process Extraction Network".

Both models have been trained with the RPROP learning algorithm \cite{riedmiller1992rprop} on the data set $\mathcal{D}$, with 70\% training and 30\% validation data for early stopping.
The training process was repeated 8 times and the networks with the lowest validation error were chosen.
In our experiments, we could validate that the training process of these RNNs is stable and the results depend little on the chosen learning algorithm.

\begin{figure}[!t]
    \centering
    \includegraphics[trim={1.1cm 3.7cm 1.0cm 1.5cm},clip,width=.5\textwidth]{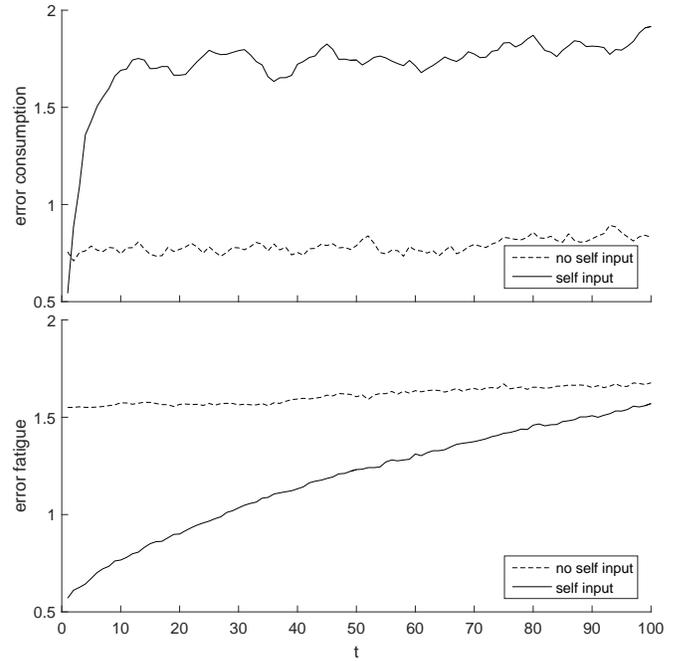}
    \caption{
    Error comparison of system model $m$. 
    We trained two different types of neural approximation models,  $m_c$ and $m_f$, for each of the reward relevant variables, consumption $c_{t+1}=m_c({o_c}_t)$ and fatigue $f_{t+1}=m_f({o_f}_t)$.
    In the first type, called 'no self input' setting, the models had to predict their respective variable without getting that variable as input in the networks, i.e.~${o_c}_t={o_f}_t=o_t\backslash \{c_t,f_t\}$, with $o_t$ the full IB observation.
    Experiences with real industrial applications have shown, that under some circumstances negative feedback effects can occur and corrupt the long term predictions in rollout settings.
    In contrast, for other learning tasks the prediction quality could be dramatically increased, if the model received a history of the variable it had to forecast, called the 'self input' setting, i.e.~${o_c}_t=o_t\backslash \{f_t\}$ and ${o_f}_t=o_t\backslash \{c_t\}$.
    For this experiment, we compared both types of RNNs by rolling out randomly drawn trajectories included in $\mathcal{D}$ on $m_c$ and $m_f$, and calculating the average absolute errors in each step $t$ with respect to the true variables' future values given by the observations in $\mathcal{D}$.
    Note that, while the network's prediction accuracy of consumption dramatically collapses after a few time steps when self input is applied, the prediction error remains almost at the same level if no self input is given to $m_c$.
    In contrast to that, the prediction of fatigue benefits from seeing historic fatigue values, at least in the evaluated period of $t<100$.
    As a result, we decided to forecast consumption without self input and fatigue with self input.
    }
    \label{m_quality}
\end{figure}

Fig. \ref{m_quality} depicts the squared error of the two selected RNNs with respect to the true IB values and describes the 'no self input' and 'self input' design of both RNNs.

\subsection{RCNN}

The Recurrent Control Neural Network (RCNN) \cite{schaefer2007recurrent} consists of two parts. 
One is a system model $m$ which is trained to predict the return by a rollout of length $T$.
The second part is a policy network, that computes for each step of the rollout an action to be fed in the system model.
This policy network takes as input the internal state of the system model $m$, which has the Markov property approximately \cite{Duell:2012b}.

The policy network has been trained with the same data set $\mathcal{D}$, as the system model $m$. 
It uses the $30 + 30$ neurons of the internal states of the consumption and the fatigue networks, $m_c$ and $m_f$, as input, followed by two hidden layers with 12 and 6 neurons, respectively, and three output neurons to encode the changes in the three steering variables, velocity, gain, and shift.
The hidden layers use hyperbolic tangent as activation function, the output layer uses the sine function.
All these configuration parameters have been chosen with almost no tuning.
Some tuning has been necessary to configure the training process of the policy network, though.
Neither for RPROP, nor Vario-Eta \cite{NeuZim:2012}, nor momentum-backprop stable training behaviors have been observed.
The best results have been observed with online-backprop with small mini-batches and small learning rates $\eta$.
We used random learning rates between $\eta=10^{-4}$ and $\eta=10^{-6}$, uniformly chosen in the logarithm of $\eta$, and a batch size of one.

\begin{figure}[!t]
    \centering
    \includegraphics[width=.5\textwidth]{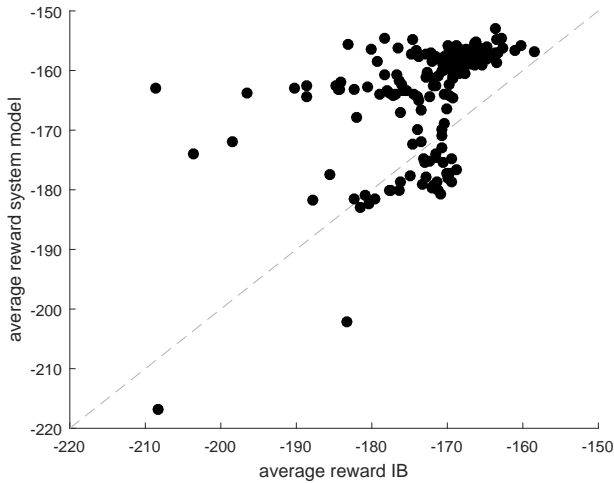}
    \caption{RCNN self assessment. 
      The average reward, estimated by the simulation model for the next 50 time steps, can be used to select good policies. 
      The plot shows the relation between average reward as estimated by the simulation model and the true average reward measured on the IB for several policies, trained by the RCNN. 
The policies were generated in 10 independent training runs and include not only the fully trained policies, but also those intermediate ones, trained for 1, 2, 4, 6, 10, 15, 23, 34, 50, 72, 104, 149, 213, 303, 431, 612, 868, 1230, 1742, 2466, 3490, and 4938 epochs, respectively. }
    \label{rcnn_self_assessment}
\end{figure}

One note concerning the possibility to assess the quality of a trained policy without executing it on the "real system" \cite{Hans:2011}, here the IB: 
if the validation error of a system model is lower on average for a rollout of sufficient length, it is the better system model (Fig. \ref{m_quality}).
If the selected system model estimates a higher return over the rollout of sufficient length for a given policy, that policy is likely to perform better, when executed on the IB (Fig. \ref{rcnn_self_assessment}).

\subsection{NFQ}

The policies of Neural Fitted Q-iteration (NFQ) \cite{riedmiller:051} were trained using a [9-20-1]-layered feed-forward MLP, with 9 neurons on the input layer for the observation $o_t$ and action $a_t$, and 20 neurons on the hidden layer. 
The output layer comprises one neuron for the associated $Q$-value $Q(o_t, a_t)$. 
All neurons of the neural network utilize a logistic activation function. 
Since NFQ is an algorithm for discrete actions, we discretized the three delta steerings towards a setting of either $-1$, $0$, or $1$ to each steering, which in total yields $\vert\mathcal{A}\vert = 27$ different actions. 

The weights of the networks were trained using non-batch stochastic gradient descent with a manually tuned constant learning rate of $\eta=0.1$.
This setting produced better results than using RPROP, as suggested by Riedmiller \cite{riedmiller:051}, because weights trained with RPROP tended to be unstable during learning on our dataset. 
Before starting the training, the $100\,000$ training samples from $\mathcal{D}$ were permuted. 
Furthermore, we divided $\mathcal{D}$ into a set of training data ($90\%$) and validation data ($10\%$) for early-stopping. 
The input data is presented to the neural network using a Z-score transformation. 
The Q-values of the output layer were scaled into the interval of the activation function as proposed by Gabel et al. \cite{GLR11}.

The overall training and evaluation procedure is as follows. 
After creating an MLP with random initial weights from the interval $[-0.1, 0.1]$, NFQ is performed over 200 iterations. 
Each row of training data is presented to the neural network for a maximum of 300 training epochs, in case the error on the validation set does not start to raise within 10 epochs. 
During the experiments we observed that the performance of NFQ policies, once successfully learned, can degrade over time.
This is consistent with findings in \cite{gordon:95,thrun:93,gabel:06}.
Therefore, we utilized a policy selection process in each experiment: we evaluated the $Q$-function after each NFQ iteration on the system model $m$ and saved the policy with the best performance according to $m$. 
Subsequently, only this policy is retained and its performance is evaluated on the IB over 10 initial states (set points $p=\{10,20,\ldots,100\}$). 

\subsection{PSO-P}

In the PSO-P setup we applied a PSO search on $m$ with 100 particles searching for 100 iterations until the best trajectory found so far was returned. 
The planning time horizon was set to $T=50$, which yielded $\gamma=0.25^{1/49}=0.9721$ as discount factor. 
The particles were arranged in the so called star topology, i.e.~each particle connected with every single other particle in the swarm \cite{eberhart:95}.

The calculation of the particles' fitnesses could be computed in parallel on 96 CPUs \footnote{Intel Xeon CPU E7-4850 v2 @ 2.30\,GHz}, resulting in an overall computation time of less than 8 seconds to compute $a_t=\pi(o_{t})$. 
While today the computation time might still be too long for several real-world industrial applications, in the future the increase in CPU speed and/or parallelization, as well as computation on GPU clusters might make PSO-P computational tractable for more and more applications.

\section{Discussion}

All of the three applied RL techniques were able to produce decent results on the IB. 
The average rewards per step are given in tables in Appendix \ref{result_tables}. 

Fig. \ref{setpointAll_compare} condenses the results of 30 RL policies and highlights the superior performance of PSO-P. 
This RL technique produces significantly more robust results than NFQ and RCNN. 
Note that all of the techniques have been trained/evaluated on the same system model $m$.

\begin{figure}[!t]
    \centering
    \includegraphics[width=.5\textwidth]{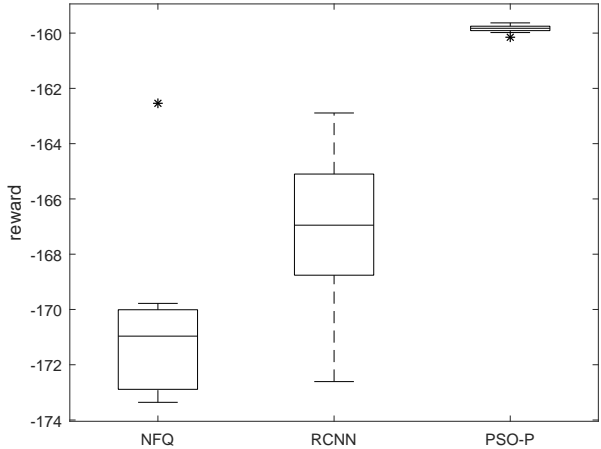}
    \caption{Comparison of NFQ, RCNN, and PSO-P experiment results. 
    Boxes cover 50\% of the data around the median (line in the box), whiskers show maximum/minimum of the data, and outliers (more than 1.5 interquantile range) are depicted as '*'. 
    On average over all set points PSO-P not only yielded the best performing policies, but also produced the most stable results, compared to NFQ and RCNN.}
    \label{setpointAll_compare}
\end{figure}

\begin{figure}[!t]
    \centering
    \includegraphics[trim={1.5cm 2.9cm 3.9cm 0.9cm},clip,width=.5\textwidth]{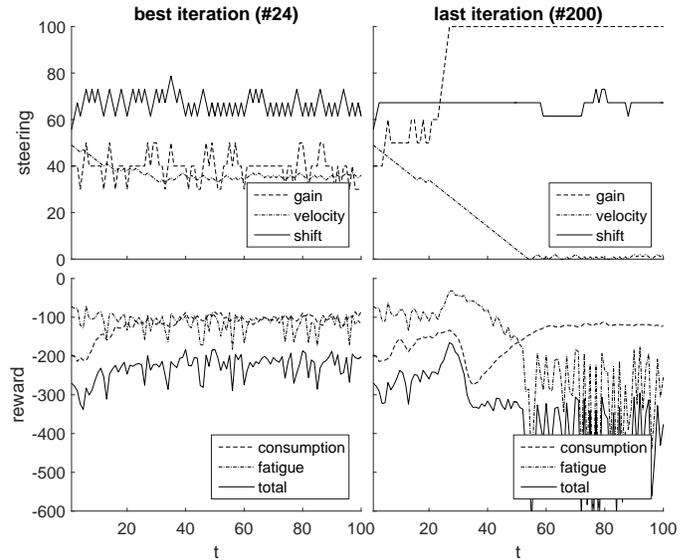}
    \caption{Comparison of NFQ trajectories from different iterations.
    During the experiments with NFQ on IB, we noticed that this technique's training process is highly unstable.
    While during the first NFQ iterations the resulting policies constantly improve their performance, this process starts decreasing after about 30 to 40 iterations.
    In later iterations the performance completely drops and the percentage of completely incapable policies increases.
    The given example in the figure above depicts two NFQ policies from different iterations out of the same experiment.
    In the first column steerings and resulting rewards are plotted from a policy selected by its performance predicted by system model $m$.
    In the second column, steerings and rewards are given from iteration 200 of the NFQ experiment.
    It is obvious that the second policy is completely incapable of steering the IB.
    }
    \label{nfq_trajectories}
\end{figure}

\begin{figure}[!t]
    \centering
    \includegraphics[trim={0.6cm 5.1cm 0.6cm 1.6cm},clip,width=.5\textwidth]{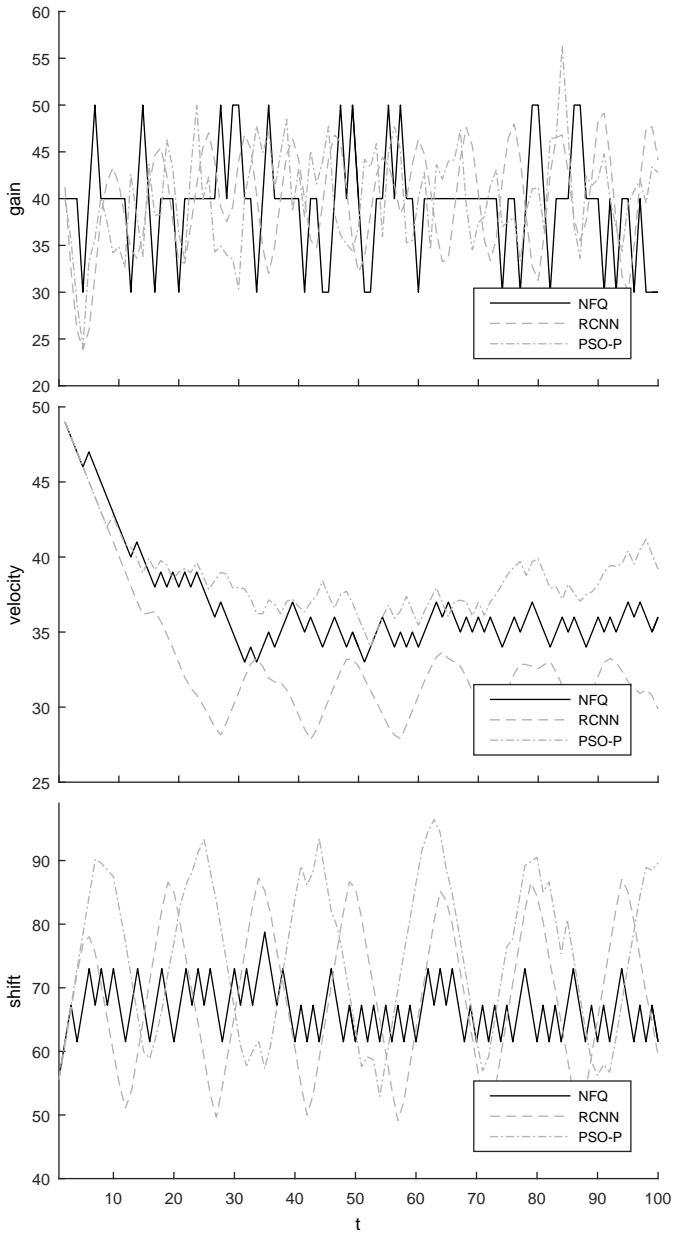}
    \caption{The best performing NFQ trajectory. 
    In comparison to the other RL techniques, NFQ operates in the same region of gain steering, between 30 and 50. 
    Similarly to PSO-P it throttles velocity down to 35. For the shift steering, NFQ could not follow the optimal strategy, which is a periodic trajectory with amplitude 20 (around 70) and cycle duration of 24. 
    It maintains shift around 70, which is indeed a local optimum, which requires no cyclic behavior.}
    \label{nfq_trajectory}
\end{figure}

The NFQ results are of the lowest performance in our experiments. 
This can be partially explained by the fact that NFQ applied only discrete actions, which is some limitation given that the IB is designed to work on continuous action spaces.
Nevertheless, a second NFQ-inherent problem which has been revealed, is its highly unstable training behavior in off-policy, batch-based trainings settings.
The training process itself gives no answer to the question when to stop the training.
One might think that it might be a good plan to perform the training as long as computationally feasible, and use the result from the last iteration.
In our experiments, this procedure would have only created bad policies (see Fig. \ref{nfq_trajectories}).
Significantly better results were achieved by evaluating the resulting NFQ policy of each iteration with the system model $m$.
The policy which produced the highest approximated average reward was then declared the best NFQ policy of each experiment.
Fig. \ref{nfq_trajectory} gives a detailed explanation on some properties of the best performing NFQ policy.

\begin{figure}[!t]
    \centering
    \includegraphics[trim={0.6cm 5.1cm 0.6cm 1.6cm},clip,width=.5\textwidth]{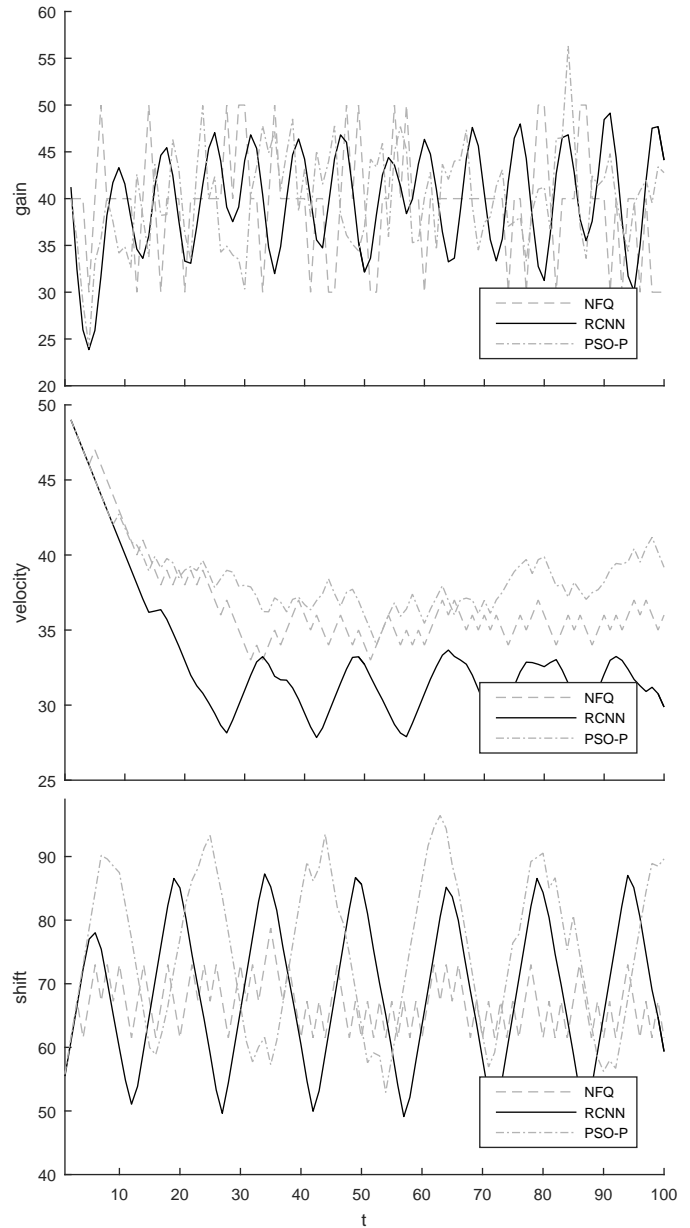}
    \caption{The best performing RCNN trajectory. 
    As closed-form policy RCNN tends to generate smoother and more regular trajectories. 
    Despite the IB is not implying a strict periodic change in gain, the depicted RCNN policy performs a very regular gain trajectory. 
    RCNN reduces velocity significantly more than NFQ and PSO-P did. 
    Again, the trajectory looks very regular. 
    Since the velocity change has the exact same cyclic duration as shift steering, this might be a cross talk like effect, caused by shared weights of both policy outputs.
    Despite RCNN discovered that it is beneficial to apply a cyclic trajectory in shift, it was not capable to reveal the true underlying cyclic duration of 24 which would have produced the highest reward.}
    \label{rcnn_trajectory}
\end{figure}

The RCNN experiments produced better results, compared to NFQ. 
RCNN policies operate in the continuous IB action space and yield compact closed-form policies.
Even though all of the trained policies yielded similar training errors, their real performance evaluated on the IB differs quite a lot.
This property implies that RCNN is rather sensitive towards prediction errors of the system model $m$. 
Fig. \ref{rcnn_trajectory} gives a detailed explanation on some properties of the best performing RCNN policy.

\begin{figure}[!t]
    \centering
    \includegraphics[trim={0.6cm 5.1cm 0.6cm 1.6cm},clip,width=.5\textwidth]{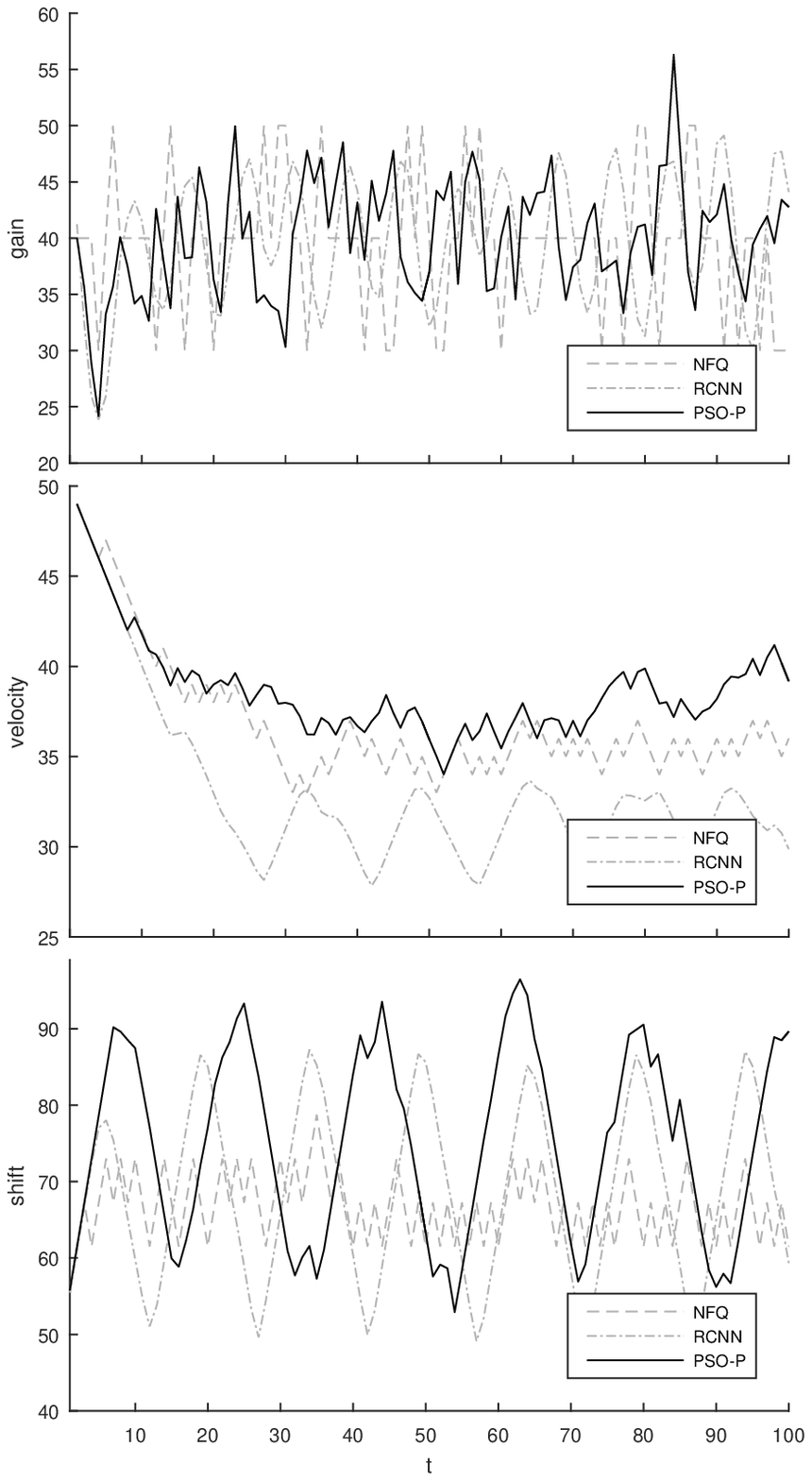}
    \caption{The best performing PSO-P trajectory. 
    Since PSO-P is not bound on a specific policy representation, it has the capability to freely change all steerings to wherever the model predicts the best reward.
    This leads to potentially less smooth and less regular trajectories.
    Similarly to NFQ and RCNN, PSO-P maintains gain around 40 (+/-10 depending on stochastically occurring effects).
    Velocity levels around 38, which is optimal for IB at set point 100.
    PSO-P has been the only technique capable of following the optimal shift strategy, which is given by a cyclic duration of 24 and an amplitude of 20 (around 70).
    Due to the initially randomly distributed particles, trajectories are prone to appear slightly irregular.
    This effect can be tackled by applying more particles, as well as more PSO iterations, if computationally feasible.}
    \label{psop_trajectory}
\end{figure}

In our experiments PSO-P has demonstrated high reward performance and outstanding robustness, that have been observed before \cite{hein:16}.
For 8 out of 10 set point values for $p$, PSO-P yielded the best RL policy on average.
Moreover, the performances of all experiments were very close to each other, which implies a high robustness against different initial PSO conditions, like particle positions and velocities.
Fig. \ref{psop_trajectory} gives a detailed explanation on some properties of the best performing PSO-P result.

\section{Conclusion}

In this paper, we have compared the new RL approach PSO-P with two standard RL techniques, RCNN and NFQ, on a recently introduced industrial benchmark. 
This benchmark has been designed to imitate realistic behavior and aspects which can be found in real-world industrial applications.

The experiments show important steps of the off-policy, batch based method stack necessary for applying RL in industrial facilities.
Starting with limited exploration data, an RNN system model is trained and tested. 
Such a model is crucial, because applying random policies on the real system is usually prohibited in real-world applications. 
Despite NFQ being classified as a model-free RL technique, our experiments show that it still requires a precise system model for policy selection.
The same model has been used for either training a closed-form neural network policy (RCNN), policy selection (NFQ), and exploiting the model for finding optimal actions (PSO-P).

NFQ, with its inherent limitation to discrete actions, and its tendency for instability during the training process, produced the worst performing policies.
Although higher performance could be achieved, for example, by increasing the discrete action space and approximating the Markov state by concatenating historic observations.

RCNN, with its ability to apply continuous actions and an inherent policy performance measure, computed closed-form policies of good performance.
Possible improvements are, for example, different network topologies, bigger neural networks, and more advanced neural learning algorithms.

PSO-P demonstrated the best performance with unmatched robustness.
Out of the box, by only setting few, easy to determine parameters, it produced the best results for almost every set point.
The biggest disadvantage of this technique lies in the computational effort required for the determination of the next action.
In our experiments the next action has been computed in less than 8 seconds, which is still too long for many industrial applications.
Improvements increasing computational power and speed might lower this value, until it becomes feasible for more and more applications.

In summary, first experiences have been made with the IB, which indeed contains many realistic objectives, issues, and features.
Experiments have shown that the benchmark could be solved by three completely different RL techniques in a realistic off-policy, batch-based setting.

\section*{Acknowledgment}

The project this report is based on was supported with funds from the German Federal Ministry of Education and Research under project number 01IB15001. The sole responsibility for the report's contents lies with the authors.

\appendices

\section{Result tables}
\label{result_tables}
Tables \ref{table:nfq_results}, \ref{table:rcnn_results}, and \ref{table:psop_results} contain the average per step rewards for each of the experiments.
The maximum achievable average reward is given in brackets in the first column.
These values have been computed by applying PSO-P on the real IB system dynamics under preservation of the initial seed of the pseudo-random number generator, i.e.~the optimizer searched for the best actions given a fixed and infinitely replicable future.
As a result, a very accurate estimate of the maximum achievable average reward for each initial Markov state has been evaluated.

\begin{table*}[!ht]
	\begin{center}
		\begin{tabular}{l|rrrrrrrrrr|r} 
    		\hline\hline
			SetPoint  (MAX) & \multicolumn{1}{c}{1} & \multicolumn{1}{c}{2} & \multicolumn{1}{c}{3} & \multicolumn{1}{c}{4} & \multicolumn{1}{c}{5} & \multicolumn{1}{c}{6} & \multicolumn{1}{c}{7} & \multicolumn{1}{c}{8} & \multicolumn{1}{c}{9} & \multicolumn{1}{c}{10} & \multicolumn{1}{|c}{mean}\\
    		\hline
			10 (-79.75) & -97.54 & -91.73 & -100.28 & -102.13 & -110.90 & -96.39 & -121.73 & -108.40 & -94.13 & -99.15 & -102.24\\
			20 (-93.58) & -113.65 & -106.15 & -115.27 & -114.54 & -115.34 & -111.69 & -117.59 & -124.10 & -108.64 & -114.33 & -114.13\\
			30 (-110.11) & -128.00 & -120.58 & -131.55 & -117.85 & -123.61 & -127.96 & -128.02 & -137.19 & -123.20 & -130.77 & -126.87\\
			40 (-126.56) & -148.69 & -143.55 & -147.77 & -146.44 & -146.80 & -146.54 & -152.92 & -152.91 & -150.42 & -152.38 & -148.84\\
			50 (-142.12) & -168.32 & -161.11 & -163.66 & -165.25 & -165.81 & -169.29 & -166.46 & -165.30 & -170.58 & -186.71 & -168.25\\
			60 (-156.31) & -189.12 & -178.16 & -183.03 & -185.21 & -184.79 & -185.52 & -177.50 & -180.39 & -189.63 & -184.33 & -183.77\\
			70 (-168.78) & -193.54 & -184.67 & -198.86 & -209.53 & -188.57 & -188.20 & -194.07 & -189.06 & -208.26 & -191.69 & -194.65\\
			80 (-180.43) & -206.33 & -196.34 & -219.95 & -212.50 & -204.29 & -201.38 & -208.81 & -203.94 & -220.41 & -211.30 & -208.53\\
			90 (-192.28) & -219.38 & -212.64 & -216.72 & -210.98 & -218.49 & -220.62 & -223.82 & -221.01 & -220.74 & -225.84 & -219.02\\
			100 (-204.96) & -235.56 & -230.50 & -236.57 & -235.94 & -239.23 & -258.04 & -242.68 & -246.58 & -242.09 & -236.06 & -240.33\\
			\hline  	
			mean (-145.49) & -170.01 & -162.54 & -171.37 & -170.04 & -169.78 & -170.56 & -173.36 & -172.89 & -172.81 & -173.26 & -170.66\\
    		\hline\hline		
    	\end{tabular}
	    \smallskip
		\caption{NFQ results.}
		\label{table:nfq_results}
	\end{center}
\end{table*}

\begin{table*}[!ht]
	\begin{center}
		\begin{tabular}{l|rrrrrrrrrr|r} 
    		\hline\hline
			SetPoint  (MAX) & \multicolumn{1}{c}{1} & \multicolumn{1}{c}{2} & \multicolumn{1}{c}{3} & \multicolumn{1}{c}{4} & \multicolumn{1}{c}{5} & \multicolumn{1}{c}{6} & \multicolumn{1}{c}{7} & \multicolumn{1}{c}{8} & \multicolumn{1}{c}{9} & \multicolumn{1}{c}{10} & \multicolumn{1}{|c}{mean}\\
    		\hline
			10 (-79.75) & -111.13 & -122.26 & -119.34 & -95.92 & -115.09 & -113.91 & -105.85 & -115.09 & -95.82 & -107.17 & -110.16\\
			20 (-93.58) & -121.90 & -130.82 &-124.06 & -109.83 & -123.14 & -123.50 & -119.57 & -126.83 & -112.09 & -116.14 & -120.79\\
			30 (-110.11) & -133.23 & -139.25 & -129.04 & -131.62 & -132.91 & -133.69 & -132.71 & -136.34 & -129.42 & -129.39 & -132.76\\
			40 (-126.56) & -146.56 & -150.49 & -144.37 & -147.49 & -147.41 & -144.81 & -146.30 & -146.29 & -150.49 & -142.14 & -146.64\\
			50 (-142.12) & -160.55 & -161.44 & -160.18 & -161.01 & -157.81 & -155.15 & -160.98 & -158.45 & -160.11 & -156.35 & -159.20\\
			60 (-156.31) & -170.94 & -171.83 & -172.42 & -175.24 & -168.43 & -167.95 & -170.20 & -171.79 & -169.21 & -167.76 & \textbf{-170.58}\\
			70 (-168.78) & -182.35 & -185.99 & -183.43 & -189.68 & -181.03 & -185.87 & -183.51 & -185.78 & -180.59 & -178.81 & \textbf{-183.70}\\
			80 (-180.43) & -193.09 & -200.26 & -199.50 & -204.72 & -194.57 & -202.84 & -200.87 & -200.30 & -194.15 & -194.15 & -198.45\\
			90 (-192.28) & -206.85 & -220.77 & -212.60 & -216.20 & -210.32 & -214.43 & -220.20 & -220.42 & -210.39 & -211.04 & -214.32\\
			100 (-204.96) & -224.36 & -242.95 & -232.14 & -232.37 & -226.57 & -232.74 & -247.43 & -238.12 & -226.63 & -232.22 & -233.55\\  
			\hline  	
			mean (-145.49) & -165.10 & -172.61 & -167.71 & -166.41 & -165.73 & -167.49 & -168.76 & -169.94 & -162.89 & -163.52 & -167.01\\
			\hline\hline		
		\end{tabular}
		\smallskip
		\caption{RCNN results.}
		\label{table:rcnn_results}
	\end{center}
\end{table*}

\begin{table*}[!ht]
	\begin{center}
		\begin{tabular}{l|rrrrrrrrrr|r} 
    		\hline\hline
			SetPoint  (MAX) & \multicolumn{1}{c}{1} & \multicolumn{1}{c}{2} & \multicolumn{1}{c}{3} & \multicolumn{1}{c}{4} & \multicolumn{1}{c}{5} & \multicolumn{1}{c}{6} & \multicolumn{1}{c}{7} & \multicolumn{1}{c}{8} & \multicolumn{1}{c}{9} & \multicolumn{1}{c}{10} & \multicolumn{1}{|c}{mean}\\
    		\hline
			10 (-79.75) & -85.97 & -89.25 & -87.58 & -88.03 & -87.97 & -87.72 & -87.59 & -88.55 & -88.01 & -87.64 & \textbf{-87.83}\\
			20 (-93.58) & -102.46 & -102.07 & -102.46 & -101.38 & -101.12 & -102.04 & -102.70 & -102.13 & -102.71 & -102.25 & \textbf{-102.13}\\
			30 (-110.11) & -118.54 & -118.13 & -119.55 & -117.55 & -117.23 & -118.27 & -118.51 & -118.39 & -117.49 & -117.26 & \textbf{-118.09}\\
			40 (-126.56) & -136.80 & -136.57 & -135.76 & -136.26 & -135.66 & -137.12 & -135.83 & -137.30 & -136.21 & -136.80 & \textbf{-136.43}\\
			50 (-142.12) & -154.10 & -156.62 & -154.99 & -155.11 & -155.77 & -155.19 & -154.13 & -155.18 & -156.82 & -154.55 & \textbf{-155.25}\\
			60 (-156.31) & -173.90 & -173.86 & -173.22 & -175.07 & -174.76 & -173.95 & -174.12 & -173.48 & -173.71 & -172.57 & -173.86\\
			70 (-168.78) & -193.00 & -192.67 & -191.84 & -191.86 & -193.89 & -193.64 & -192.05 & -191.90 & -192.69 & -192.51 & -192.61\\
			80 (-180.43) & -196.07 & -195.89 & -199.28 & -197.75 & -196.35 & -195.48 & -197.76 & -195.66 & -196.52 & -199.10 & \textbf{-196.99}\\
			90 (-192.28) & -209.17 & -208.94 & -209.24 & -210.24 & -210.77 & -208.36 & -208.99 & -209.33 & -209.46 & -208.96 & \textbf{-209.35}\\
			100 (-204.96) & -226.24 & -227.52 & -225.14 & -225.00 & -226.25 & -226.45 & -224.84 & -227.15 & -223.88 & -226.27 & \textbf{-225.87}\\
			\hline  	
			mean (-145.49) & -159.63 & -160.15 & -159.91 & -159.83 & -159.98 & -159.82 & -159.65 & -159.91 & -159.75 &-159.79 & \textbf{-159.84}\\
    		\hline\hline		
    	\end{tabular}
	    \smallskip
		\caption{PSO-P results.}
		\label{table:psop_results}
	\end{center}
\end{table*}

\bibliographystyle{IEEEtran}
\bibliography{bibliography}

\end{document}